%% file: T1.tex
\documentclass[technote,11pt,draftclsnofoot,onecolumn]{IEEEtran}

\usepackage{times}
\usepackage{epsfig}
\usepackage{subfig}
\usepackage{graphicx}
\usepackage{amsmath}
\usepackage{amssymb}
\usepackage{wrapfig}

\newtheorem{theorem}{Theorem}[section]

\newtheorem{lemma}[theorem]{Lemma}
\newtheorem{proposition}[theorem]{Proposition}

\newenvironment{proof}[1][Proof]{\begin{trivlist}
\item[\hskip \labelsep {\bfseries #1}]}{\end{trivlist}}

\newcommand{\qed}{\nobreak \ifvmode \relax \else
      \ifdim\lastskip<1.5em \hskip-\lastskip
      \hskip1.5em plus0em minus0.5em \fi \nobreak
      \vrule height0.75em width0.5em depth0.25em\fi}

\newtheorem{assumption}[theorem]{Assumption}

\begin{document}
\title{The Stability of Convergence of Curve Evolutions in Vector Fields}
\author{Junyan Wang*~\IEEEmembership{Member, IEEE} and Kap Luk Chan~\IEEEmembership{Member, IEEE}
\thanks{Junyan Wang and Kap Luk Chan are with the
School of Electrical and Electronic Engineering, Nanyang Technological University, Singapore 639798 e-mail: \{wa0009an,eklcan\}@ ntu.edu.sg}}
\markboth{}%
{}
\maketitle

\begin{abstract}
\boldmath
Curve evolution is often used to solve computer vision problems. If the curve evolution fails to converge, we would not be able to solve the targeted problem in a lifetime. This paper studies the theoretical aspect of the convergence of a type of general curve evolutions. We establish a theory for analyzing and improving the stability of the convergence of the general curve evolutions. Based on this theory, we ascertain that the convergence of a known curve evolution is marginal stable. We propose a way of modifying the original curve evolution equation to improve the stability of the convergence according to our theory. Numerical experiments show that the modification improves the convergence of the curve evolution, which validates our theory.
\end{abstract}

\begin{IEEEkeywords}
Object segmentation, active contour, geometry of curve evolution, stability of convergence.
\end{IEEEkeywords}

%
\IEEEpeerreviewmaketitle

\input{body_single}

{\small
\bibliographystyle{IEEEtran}
\bibliography{LevelSetActiveContours}
}


\ifCLASSOPTIONcaptionsoff
  \newpage
\fi

\end{document}

%% file: body_single.tex
\section{Introduction}

Various computer vision tasks have been formulated and solved based on the formulation of the curve evolution equations, such as boundary detection \cite{kass88snakes}, shape-from-shading \cite{Kimmel95LevelSetSFS} and optical flow \cite{Kumar96opticalflow}. In boundary detection, for example, the desired object region are supposed to be enclosed by the \emph{convergent} curve produced by the curve evolution. The requirement of convergence is very necessary. Suppose that the curve evolution cannot converge, the evolving curve may eventually disappear or we would never obtain a solution in a lifetime.

The curve evolutions obtained by gradient descent based functional energy minimization \cite{kass88snakes} \cite{caselles97GAC} \cite{ChanVese01ActiveCon} are \emph{globally convergent} in theory \cite{BookPrincOptDes}. Furthermore, the numerical convergence of some of those curve evolutions was also shown by simulation \cite{Chaudhury08StabConvLS} and theoretical analysis \cite{Verd08ConvAC}. Hence, the curve evolutions in this framework are convergent once formulated. Beside of the methodology of energy minimization, people have also tried to directly formulate the object boundaries as the solution to ad-hoc curve evolution equations, such as in \cite{Balloon91} \cite{Malladi95ShMo} \cite{Caselles93GeoModel} \cite{Xie08MAC} \cite{WangChanWang08PSP}. However, no general technique for either determining or ensuring the convergence of such curve evolutions is known.

As a main contribution of this paper, a theory for analyzing and improving the convergence of curve evolutions has been established. First, a condition for determining the stability of the convergence of a general curve evolution in a vector field $\vec{F}$ is derived. There could be various ways to improving the stability of a curve evolution. We also expect the improved curve evolution to behave similar to the original curve evolution. Then, an error bound that was extensively used in control theory is also adopted for measuring the similarity of curve evolutions. The similar curve evolutions are supposed to have similar stationary solution. Finally, we show the marginal stability of the convergence of the curve evolution of Equilibrium Flow proposed in \cite{WangChanWang08PSP}. We also modify the original curve evolution equation of Equilibrium Flow to improve the stability of the convergence.

The rest of the paper is organized as follows. In section \ref{SEC:BG}, we introduce background of curve evolution. In section \ref{SEC:THEORY}, we present the mathematical analysis on the convergence of curve evolution in a vector field. In section \ref{SEC:EG}, we present the application of the mathematical analysis to a system of curve evolutions. In section \ref{SEC:Exp}, experimental results validates our theoretical claims. We conclude the paper in section \ref{SEC:CON}.

\section{Background}\label{SEC:BG}
In this section, we provide the background to facilitate the subsequent discussions in this section. A general formulation of curve evolution can be written as follows \cite{OsherBookDynamic} \cite{SapiroBook}:
\begin{equation}\label{EQ:CE_TN}
\begin{split}
{\partial C(p,t)\over\partial t} &= \alpha(p,t)\vec{T}(p,t)+\beta(p,t)\vec{N}(p,t)\\
&= \alpha(p,t)\vec{T}(p,t) + \langle\vec{F},\vec{N}(p,t)\rangle\vec{N}(p,t),
\end{split}
\end{equation}
where $\alpha,\beta$ are the general form of coefficients defined along the curve, $\vec{F}(p,t)$ is the underlying vector field which is a vector function on the 2D (image) domain, $C(p,t)$ denotes the curve parameterized by $p$ evolving with respect to time $t$, $\vec{T}$ is the tangent of the curve, $\vec{N}$ is the outward normal. Throughout this paper, we discuss about the curve evolution under the assumption below.
\begin{assumption}
The vector field $\vec{F}$ only depends on the location of $C(p,t)$, i.e., $\vec{F}(p,t)=\vec{F}(\vec{x})|_{\vec{x}\in C(p,t)}$.
\end{assumption}
This implies that the curve evolution is only due to the static vector/velocity field $\vec{F}$ on the domain.

A fundamental property of the curve evolution of Equation (\ref{EQ:CE_TN}) was reported by Epstein and Gage in \cite{Epstein87CS}, known as the Lemma of curve evolution.
\begin{lemma}[Epstein and Gage]\label{LM:CE}
Given the closed curve $C(\tilde{p},t)$ parameterized by arbitrary $\tilde{p}\in \tilde{B}$ at an artificial time $t$ with the normal $\vec{N}$, the tangent $\vec{T}$ of the curve, and given the geometric flow of a curve evolution by
\begin{equation}\label{EQ:CE1}
{\partial C(\tilde{p},t)\over \partial t}=\alpha(\tilde{p},t) {\vec{T}}(C(\tilde{p},t)) +
\beta(\tilde{p},t) {\vec{N}}(C(\tilde{p},t)).
\end{equation}
If $\beta$ does not depend on the parametrization, meaning that is a geometric intrinsic characteristic of the curve, then the image of $C(\tilde{p},t)$ that satisfies Equation(\ref{EQ:CE1}) is identical to the image of the family of curves $C(p,t)$, parameterized by $p\in B$, that satisfies
\begin{equation}\label{EQ:CE2}
{\partial C(p,t)\over \partial t}= \beta(p,t)
{\vec{N}}(p,t).
\end{equation}
\end{lemma}
This can be verified by the well-known level set methods.


\section{The mathematical analysis of the convergence of curve evolution}\label{SEC:THEORY}
The results of this section are obtained based on the Lemma of curve evolution \ref{LM:CE} and the idea of using the zero level set of a signed distance function to represent closed curves. The proofs can be found in the appendix.

\begin{lemma}\label{TH:CEChain}
Given the curve evolution $C(p,t)$ defined as follows,
\begin{equation}
\begin{split}
{\partial C\over\partial t}=\epsilon(p,t)\vec{N}(p,t),
\end{split}
\end{equation}
the following holds
\begin{equation}\label{EQ:CEChain}
{\partial^{(k+1)} C\over\partial t^{(k+1)}}={\partial^k\epsilon(p,t)\over\partial t^k}\vec{N}(p,t).
\end{equation}
\end{lemma}

To prove the above lemma, we need the lemma below. The proof of Lemma \ref{TH:CEChain} is straightforward by following Lemma \ref{LM:DNN=0} and the chain rule of differentiation.
\begin{lemma}\label{LM:DNN=0}
\begin{equation}\label{EQ:DNN=0}
{d\vec{N}\over dC}(C)\vec{N}(C)=0.
\end{equation}
\end{lemma}

The above results eventually lead us to a more practical result that is stated as follows, which can be proved by substituting Equation (\ref{EQ:CEChain}) into the Taylor series.
\begin{lemma}\label{TH:DCE}
If there exists $R>0$, such that $\forall |\tau|<R$, such that the Taylor series of $C(p,t+\tau)$ at $C(p,t)$ exists, then there exists a unique $\xi(p,\tau)$, such that,
\begin{equation}\label{EQ:DCE}
C(p,t+\tau)=C(p,t)+\tau\xi(p,\tau)\vec{N}(p,t),
\end{equation}
where
\begin{equation}
\xi(p,\tau)=\sum\limits_{k=1}^{\infty}{\tau^{k-1}\over k!}{\partial^{(k-1)} \epsilon(p,t)\over\partial t^{(k-1)}}.
\end{equation}
\end{lemma}
By this lemma, we convert the curve evolution in the form of PDE to an algebraic equation to facilitate the analysis of curve evolutions in section \ref{SEC:EG}.

%
%

We now analyze the convergence of the curve evolutions. The objective is to ascertain whether the convergence is stable. Naturally, we consider the case of unstable as a lack of convergence. Given a curve evolution defined by the following,
\begin{equation}
{\partial C(p,t)\over \partial t}=\langle\vec{F},\vec{N}\rangle\vec{N}.
\end{equation}
The convergence of the curve evolution is stated as follows:
\begin{equation}
\lim\limits_{t\rightarrow \tau}{\partial C(p,t)\over \partial t}=\langle\vec{F},\vec{N}\rangle\vec{N}=0.
\end{equation}

The condition of convergence can be found by considering a small perturbation on the converged curve. If the perturbation finally vanishes at that converged curve, the curve evolution is stably convergent otherwise it is unstable. We now examine the convergence of the curve evolutions following this idea.

The curve evolution near the converged curve $C^*$ can be represented as follows.
\begin{equation}\label{EQ:STB0}
C^*=C(p,t^*)=\left.C(p,t)\right|_{t=t^*-\tau}+\eta(p,\tau),
\end{equation}
where $\eta$ is a general formulation of perturbation, $\tau=t^*-t>0$ and is small enough and monotonically \emph{decreasing} during the evolution. The fundamental condition of convergence is shown below.
\begin{theorem}\label{TH:STB}
If $J_{nn}(C^*)<0$, the curve evolution is \emph{\textbf{stable}};If $J_{nn}(C^*)=0$, the curve evolution is \emph{\textbf{marginally stable}};If $J_{nn}(C^*)>0$ the curve evolution is \emph{\textbf{unstable}}, where $J_{nn}(C^*)=[\vec{N}^TJ[\vec{F}]\vec{N}](C^*)$, $J[\vec{F}]$ is the Jacobian of vector field $\vec{F}$.
\end{theorem}
Accordingly, we can use $J[\vec{F}](C^*)$ to examine (the stability of) the convergence of curve evolution in vector field. The proof, which makes use of Lemma \ref{LM:DNN=0} extensively, is deferred to the appendix due to its length.

Suppose that the original curve evolution is unstable or marginally stable, a modification on the formulation of the curve evolution equation might be performed. By using the above theorem, it can be verified whether the modified curve evolution is stably converged. However, this leads to another question that whether the modified curve evolution maintains the function of the original one. The following theorem could be a fundamental answer to this question.
\begin{theorem}\label{TH:Bound}
For a fixed $p$, such that ${\partial S(p,t)\over\partial t}={\partial C(p,t)\over\partial t}+\vec{G}(p,t)$, and ${\partial C(p,t)\over\partial t}$ is Lipschitz continuous with the Lipschitz constant $L$, Given that $\|\vec{G}(p,t)\|\leq\mu$,$\mu>0$, then the following holds,
\begin{equation}
\begin{split}
&\|C(p,t)-S(p,t)\|\leq\|C(p,0)-S(p,0)\|e^{L(t-t_0)}\\
&~~~~~~~~~~~~~+{\mu\over L}\{e^{L(t-t_0)}-1\}
\end{split}
\end{equation}
\end{theorem}

The above theorem provides an upper bound of the difference between curve evolutions. If the difference is small, the curve evolutions will behave analogously, and we suppose the stationary solutions of them are also similar. We shall not include the proof here, since the proof of a more detailed and general bound for ordinary differential equations, related to Gronwall-Bellman inequality, is clearly stated in Chapter 3 of \cite{Khalil02Nonlinear}.

\section{Improving the convergence of GeoSnakes}\label{SEC:EG}
\subsection{Active contours with a tangential component}
In \cite{WangChanWang08PSP}, it has been noticed that the full gradients of the edge indicator, e.g. the magnitude of image gradient, along the boundaries can be uniformly small, i.e. both the normal and tangential components of the gradients of the edge indicator along the boundary are close to zero. However, the conventional active contours, such as \cite{caselles97GAC}, considers only the normal component of the gradient of the edge indicator. The curve evolution based only on the normal component often converges at the places where the tangential component is still significant. Therefore, a new active contour model termed the Geodesic Snakes (GeoSnakes) was proposed. The Euler-Lagrange (EL) equation of the GeoSnakes requires the full gradient of the edge indicator, i.e. both the normal and tangential components, to vanish at the contour of solution. Nevertheless, the direct curve evolution to solve the EL equation only solves the normal component of the EL equation, and the curve evolution can converge when the tangential component is still significant, which we call the Pseudo Stationary Phenomenon. We propose an auxiliary curve evolution equation, termed the Equilibrium flow (EF), to solve specifically for the tangential component. The full EL equation involving both the normal and tangential components is solved by alternating the direct curve evolution and the curve evolution of EF. The system of the alternating curve evolution equations is as follows.
\begin{subequations}\label{EQS:ALT}
\begin{align}
&{\partial_t C_1^k} = \langle\vec{F},\vec{N}\rangle\vec{N}\vec{N},~~C_1^k(\tau,0)=C_2^{k-1}(\tau,\infty) \label{EQ:GVFEFa} \\
&{\partial_t C_2^k} = \langle \mathcal{R}\vec{F},\vec{N}\rangle\vec{N},~~C_2^k(\tau,0)=C_1^k(\tau,\infty)\label{EQ:GVFEFb},
\end{align}
\end{subequations}
where $k = 1,2,\ldots,C_1^0(\tau,0) = C_2^0(\tau,\infty) = C_0$, $C(\tau,\infty)=\lim\limits_{k\rightarrow\infty}C_1^k(\tau,\infty)=\lim\limits_{k\rightarrow\infty}C_2^k(\tau,\infty)$ and
$\mathcal{R}$ is a rotation matrix of $2\times2$, $\vec{F}=\nabla g$ is the gradient field of a scaler field $g$. Equation (\ref{EQ:GVFEFa}) is the original gradient descent flow, Equation (\ref{EQ:GVFEFb}) is the proposed auxiliary flow called Equilibrium Flow. Note that we omit the curvature term that imposes smoothness to simplify our discussions.

\subsection{Application of the theory to GeoSnakes}

In this subsection, we apply the theoretical results obtained above to the system of (\ref{EQS:ALT}) to show that the system of curve evolutions may not converge stably. We also show that the convergence can be improved by modifying the curve evolution equation in accordance with our analysis.

\textbf{For the gradient descent flow in Equation (\ref{EQ:GVFEFa}) under $\vec{F}=\nabla g$}, where
\begin{equation}\label{EQ:STB8}
\begin{split}
J_{nn}(C^*) &= \langle \nabla^2 g\vec{N}(C^*),\vec{N}(C^*)\rangle,\\
\end{split}
\end{equation}
where $\nabla^2 g$ is the Hessian of $g$. We may use the following lemma for analyzing $J_{nn}(C^*)$.
\begin{lemma}\label{LM:OPT}
\textbf{\emph{Given}} that the curve evolution $C(p,t)$ defined by Equation (\ref{EQ:GVFEFa}) near the converged curve $C(p,t^*)$, i.e. $C(p,t^*-\delta t)$ can be expanded by Taylor series at $t^*$, within a radius of convergence $|\delta t|<R$, where $t^*$ is the time of convergence, \textbf{\emph{and given}} that the initial curve $C(p,0)$ is not on minima of $g$ (maxima of $-g$), \textbf{\emph{then}}
\begin{equation}
\langle \nabla^2 g\vec{N},\vec{N}\rangle\leq0.
\end{equation}
\end{lemma}
The proof which makes use of Theorem \ref{TH:DCE} can be found in the appendix.

%
%
%
The above result shows that $J_{nn}\leq0$ on $C^*$, which also proves the following.
\begin{proposition}\label{TH:GDSTB}
The curve evolution under $\vec{F}=\nabla g$ is stable or the worst marginally stable.
\end{proposition}

\textbf{For the EF defined by Equation (\ref{EQ:GVFEFb}) under $\mathcal{R}\vec{F}$}, where
\begin{equation}\label{EQ:STB9}
\begin{split}
J_{nn}(C^*) &= \langle \nabla^2 g\vec{N}(C^*),\vec{T}(C^*)\rangle,
\end{split}
\end{equation}
we have the following equality at $C^*$:
\begin{equation}\label{EQ:STB10}
\begin{split}
\langle\nabla g,\vec{N}\rangle=\langle\nabla g,\pm{\nabla g\over\|\nabla g\|}\rangle=\pm\|\nabla g\|.
\end{split}
\end{equation}
This is because the gradient vanishes in the tangential direction. Hence the gradient is in the normal direction. We now differentiate the above to obtain the following,
\begin{equation}\label{EQ:STB11}
\begin{split}
{d\over ds}\pm\|\nabla g(C(s))\|=\pm{\nabla g^T\over\|\nabla g\|}\nabla^2 g C_s=J_{nn},
\end{split}
\end{equation}
which proves the following.

\begin{proposition}\label{TH:EFSTB}
If we normalize the gradient field, e.g., $\|\nabla g\|=1$, then $J_{nn}(C^*)=0$, hence, the EF is marginally stable.
\end{proposition}

Let us consider the following modification of the Equilibrium Flow define by Equation (\ref{EQ:GVFEFb}):
\begin{equation}\label{EQ:EF_PT}
{\partial C(p,t)\over \partial t} = \langle \mathcal{R}^T\nabla g+\epsilon\nabla g,\vec{N}\rangle\vec{N},
\end{equation}
where $\epsilon>0$ is chosen to be a constant. It can be shown that the above modified Equilibrium Flow have better property of convergence by the follows.
\begin{proposition}
By normalizing $\|\nabla g\|$, e.g. $\|\nabla g\|=1$, the curve evolution defined by Equation(\ref{EQ:EF_PT}) is stable or the worst marginally stable.
\end{proposition}
The proof is straightforward based on the previous discussions.


We can also show that the slightly revised Equilibrium Flow (\ref{EQ:EF_PT}) will not lose its function by showing that Equation (\ref{EQ:EF_PT}) behaves analogous to the Equilibrium Flow Equation (\ref{EQ:GVFEFb}) using Theorem \ref{TH:Bound} in which $\vec{G}=\epsilon\langle\nabla g,\vec{N}\rangle\vec{N}$ and $\epsilon$ can be small. Therefore, a curve evolution  behaving analogously to the Equilibrium Flow but converging more stably is obtained.
%

\section{Experimental validations}\label{SEC:Exp}
In this section, some numerical experiments are conducted to validate the theory and to "visualize" the case that the curves can disappear, due to failing to converge before disappear, by the system of Equations (\ref{EQ:GVFEFa}) and (\ref{EQ:GVFEFb}). We also show that this is avoided by replacing Equation (\ref{EQ:GVFEFb}) with Equation (\ref{EQ:EF_PT}). The images are smoothed and the vector fields are extended and smoothed by the method of Gradient Vector Field (GVF) \cite{Xu98GVF} \cite{GVFGAC04}. We set $\epsilon=0.1$ in (\ref{EQ:EF_PT}) in all our experiments for validation of the theoretical claims. During the implementation of the system of curve evolution equations, each switch is performed if the curve length remains almost unchanged, and we run (\ref{EQ:GVFEFa}), (\ref{EQ:GVFEFb}) or (\ref{EQ:EF_PT}), and then (\ref{EQ:GVFEFa}).

\begin{figure}
\vspace{-10pt}
  \centering
 \subfloat{\includegraphics[width=0.11\textwidth]{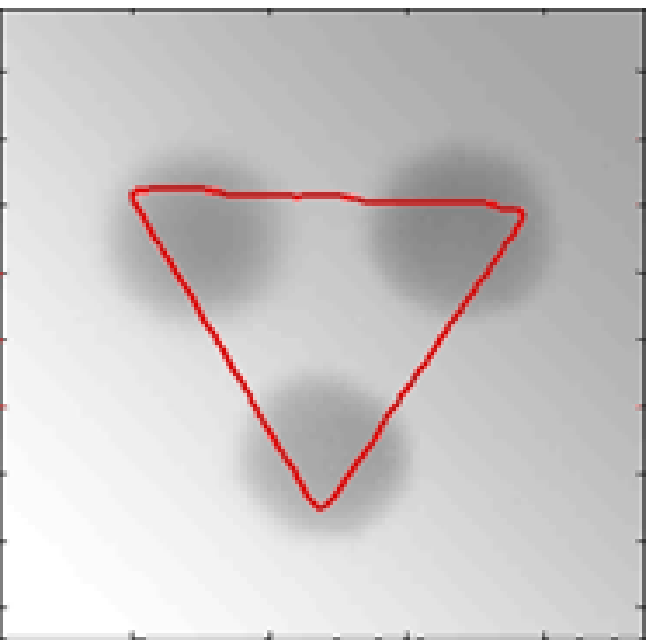}}
  \subfloat{\includegraphics[width=0.11\textwidth]{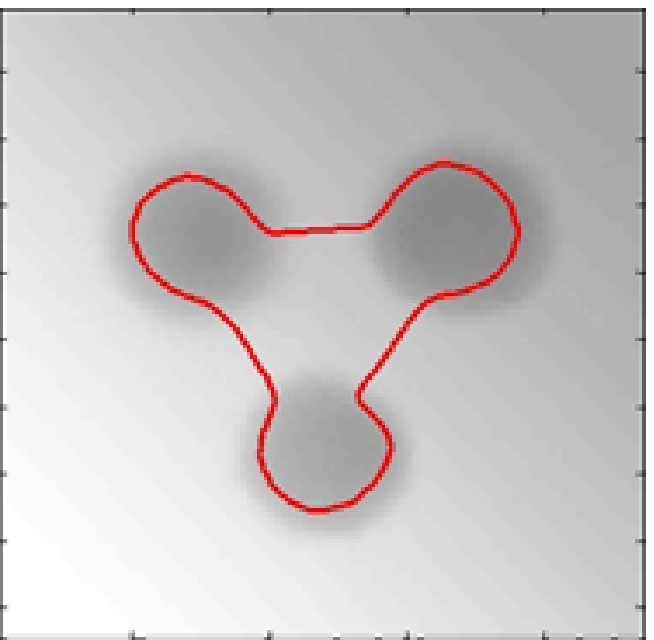}}
%
  \subfloat{\includegraphics[width=0.11\textwidth]{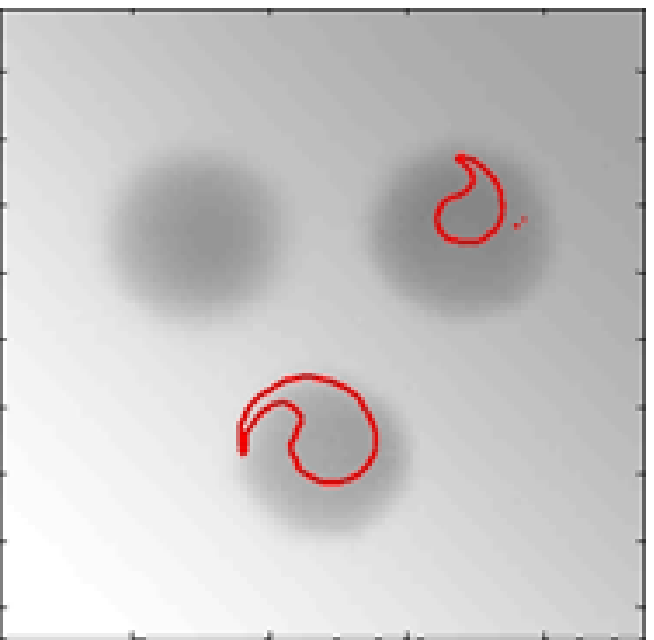}}
  \subfloat{\includegraphics[width=0.11\textwidth]{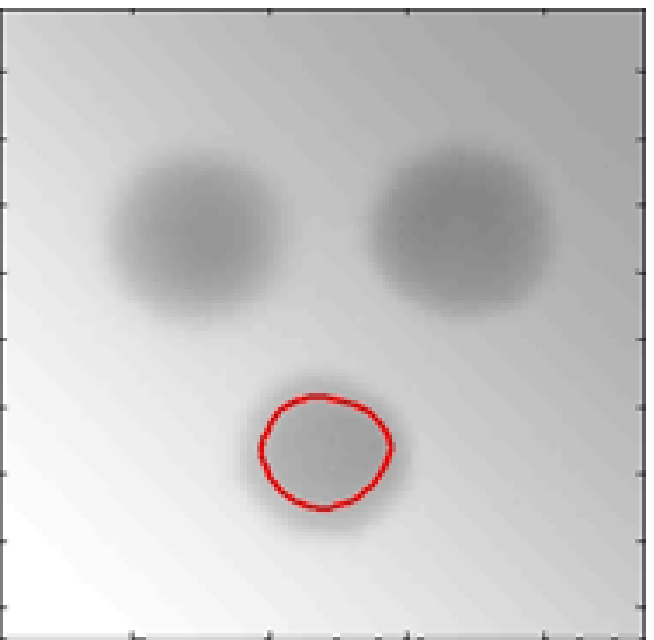}} \caption{The original framework for the pattern.}\label{Fig:Synb0}
  \subfloat{\includegraphics[width=0.11\textwidth]{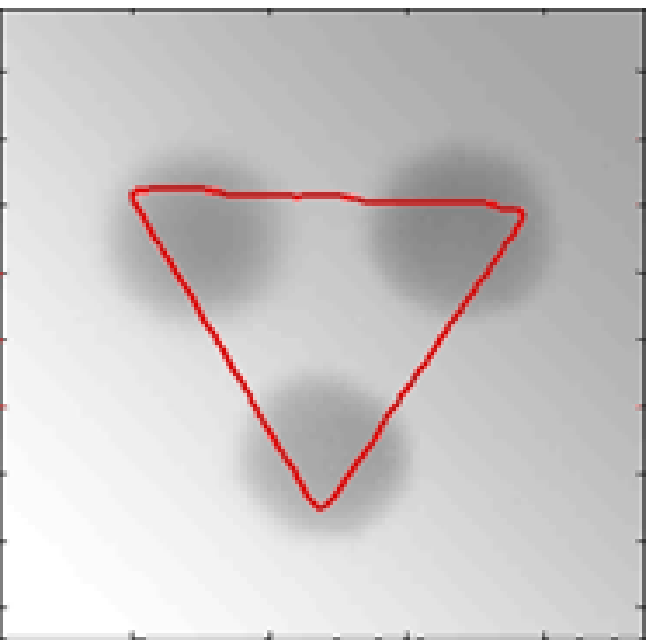}}
  \subfloat{\includegraphics[width=0.11\textwidth]{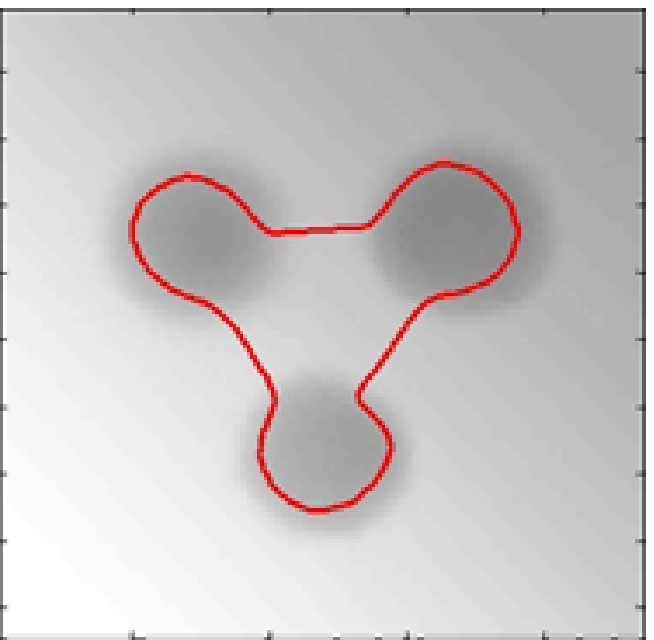}}
%
  \subfloat{\includegraphics[width=0.11\textwidth]{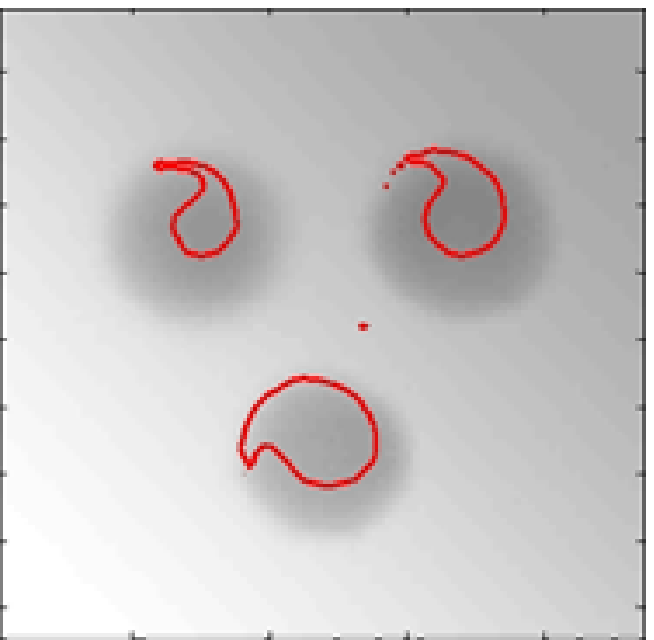}}
  \subfloat{\includegraphics[width=0.11\textwidth]{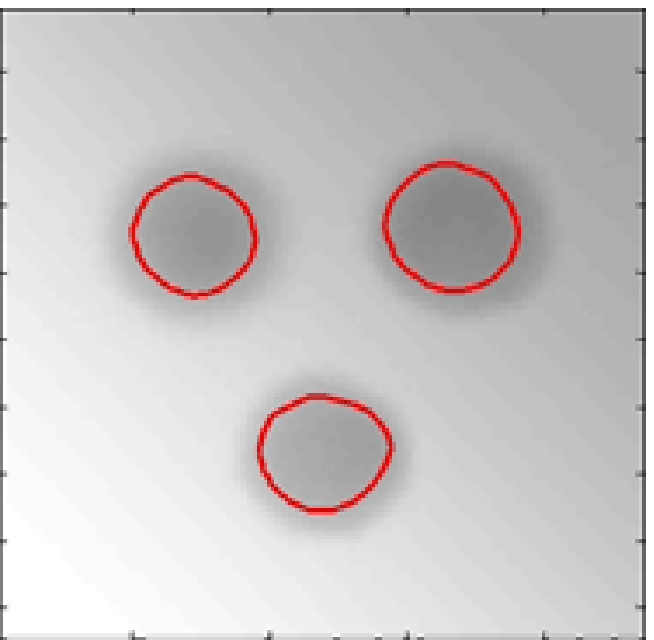}} \caption{The modified framework for the pattern.}\label{Fig:Synb1}
  \subfloat{\includegraphics[width=0.11\textwidth]{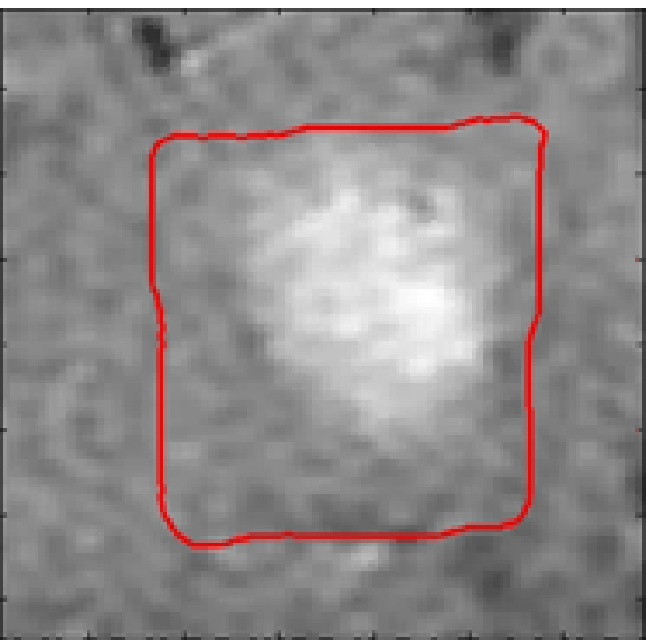}}
  \subfloat{\includegraphics[width=0.11\textwidth]{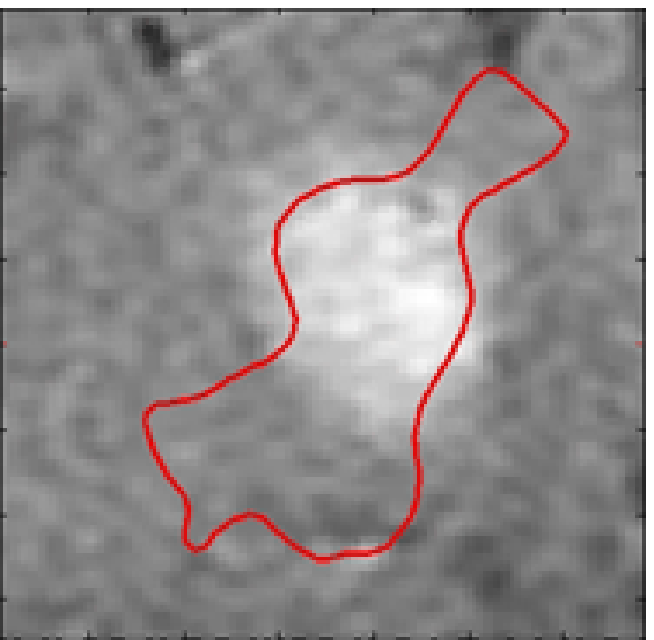}}
%
  \subfloat{\includegraphics[width=0.11\textwidth]{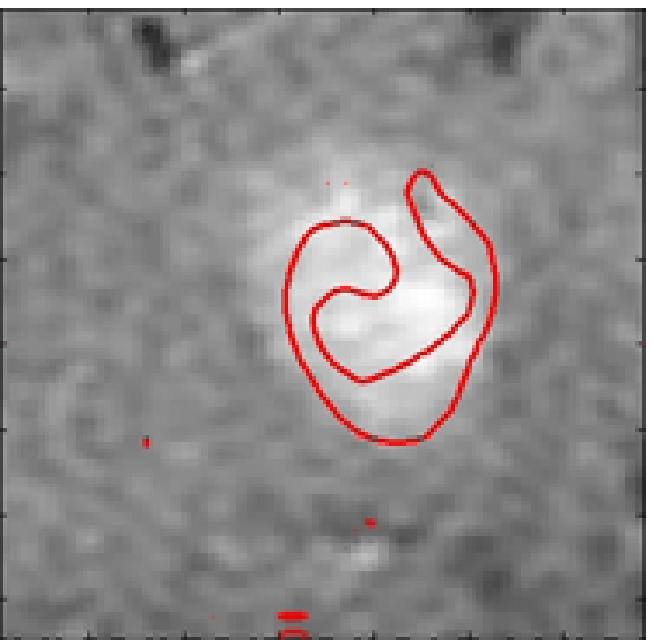}}
  \subfloat{\includegraphics[width=0.11\textwidth]{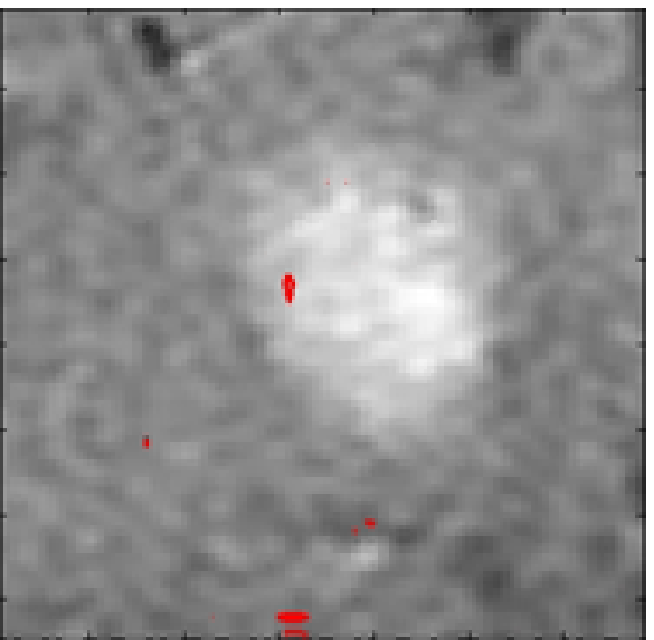}}
  \caption{The original framework for real data.}\label{Fig:Real0}
  \subfloat{\includegraphics[width=0.11\textwidth]{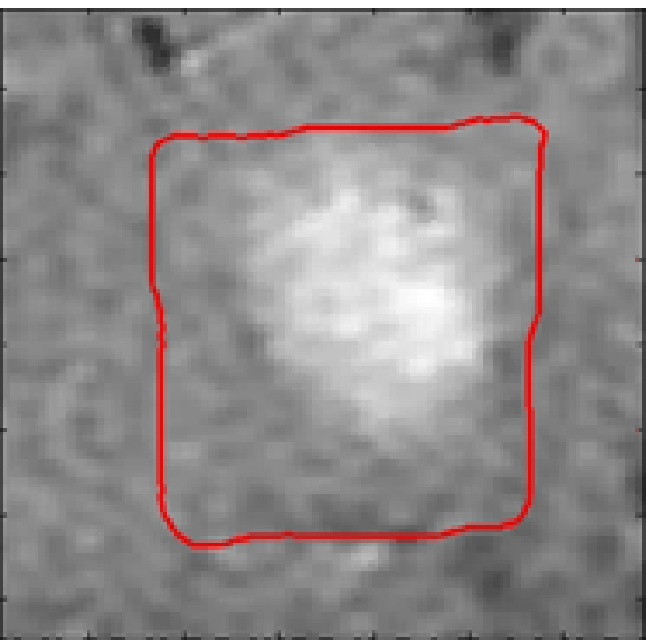}}
  \subfloat{\includegraphics[width=0.11\textwidth]{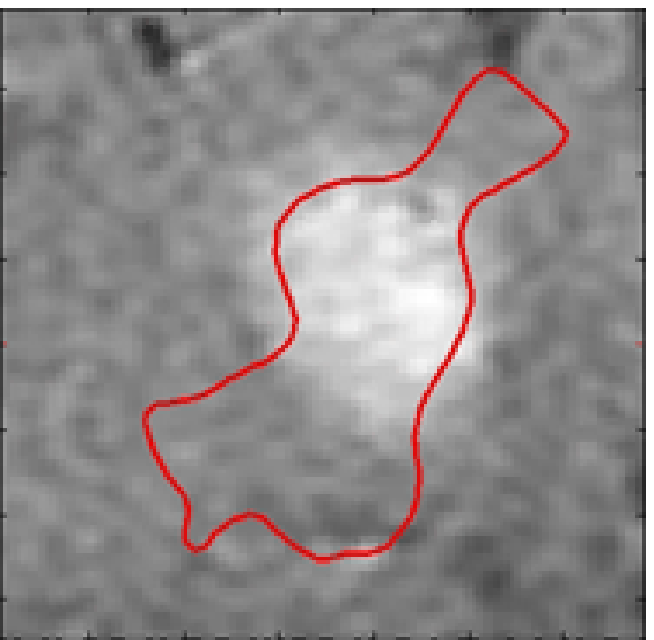}}
%
  \subfloat{\includegraphics[width=0.11\textwidth]{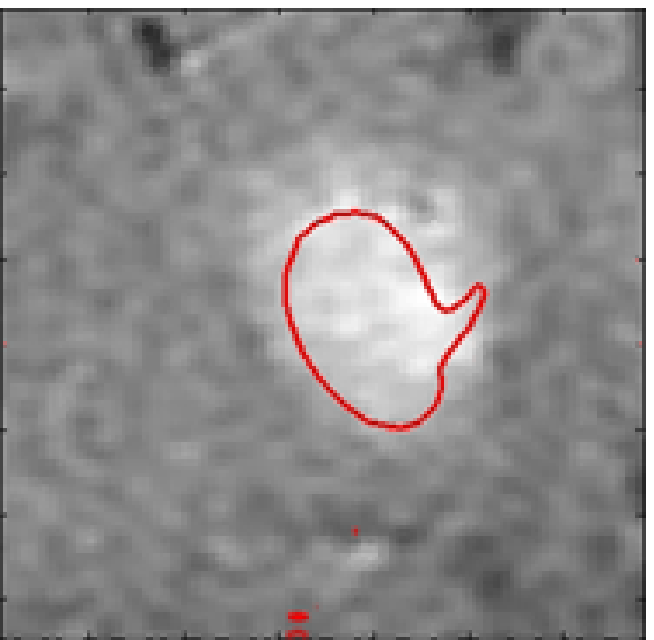}}
  \subfloat{\includegraphics[width=0.11\textwidth]{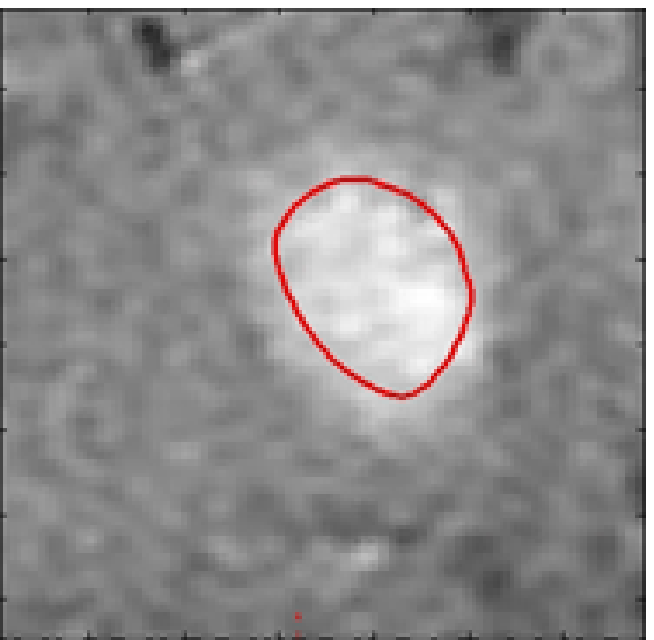}}
  \caption{The modified framework for real data.}\label{Fig:Real1}
  \vspace{-20pt}
\end{figure}

We designed a synthesized pattern, shown in Figures \ref{Fig:Synb0} and \ref{Fig:Synb1}. In this pattern, there are three blurred disks to be segmented out. The intensity in the background gradually changes. In other words, there are gradients everywhere in the background that may disturb the marginally stable Equilibrium Flow. The experiment shown in Figure \ref{Fig:Synb0} validated the theory that some of the curves for enclosing the disks finally disappear while the other disks are captured since the curve evolution is not unstable. Contrasting the lack of convergence of the original Equilibrium Flow, the improvements by the modified framework involving Equation (\ref{EQ:EF_PT}) is also shown in Figure \ref{Fig:Synb1}. The real data shown in Figures \ref{Fig:Real0} and \ref{Fig:Real1} is blur and there is gradient inside and outside the object region. We also observe the lack of convergence of the original framework in Figure \ref{Fig:Real0} as well as the improvements by the modified framework in Figure \ref{Fig:Real1}. Figures \ref{Fig:Synb0}, \ref{Fig:Real0}, \ref{Fig:Synb1} and \ref{Fig:Real1}, from the left to right, show the initialization, the converged solution by (\ref{EQ:GVFEFa}), the curve evolution by (\ref{EQ:GVFEFb}) and the final result.

\section{Discussions and conclusion}\label{SEC:CON}
The contribution of this paper is the theory on examining the stability of
convergence of the curve evolution equations formulated beforehand. The presented modification of the equilibrium flow involves an parameter $\epsilon$. It is still an open problem for determining the parameters of curve evolution equation. The presented theory is complete without choosing the specific value of the parameter, and the experiment with the pre-specified $\epsilon$ has validated the theory.

In this paper, a theory for analyzing and improving the convergence of curve evolution was established. The theory was also applied to a recently proposed framework of curve evolution to ascertain the lack of convergence of the corresponding curve evolution and to improve its stability of convergence. More applications of the developed theoretical results can be expected.

\appendices
\renewcommand{\theequation}{A-\arabic{equation}}
\setcounter{equation}{0}  
\renewcommand*\thesection{\Alph{section}}
\setcounter{section}{0}

\section*{Appendix}
\begin{proof}[Proof of Lemma \ref{LM:DNN=0}]
Firstly, we write ${d\vec{N}\over dC}(C)$ more explicitly as follows,
\begin{equation}\label{EQ:STB_NC1}
{d\vec{N}\over dC}(C)=\left(
                     \begin{array}{cc}
                       {d N^x\over dx}(C) & {d N^y\over dx}(C) \\
                       {d N^x\over dy}(C) & {d N^y\over dy}(C) \\
                     \end{array}
                   \right).
\end{equation}
Using the relation ${\nabla\Phi\over\|\nabla\Phi\|}=\vec{N}$ and $\|\nabla\Phi\|=1$ for signed distance function $\Phi$ defined by $C=\{x,y|\Phi(x,y)=0\}$, we obtain the following:
\begin{equation}\label{EQ:STB_NC2}
\nabla^2\Phi={d\vec{N}\over dC}(C),
\end{equation}
i.e., ${d\vec{N}\over dC}(C)$ is the Hessian matrix of $\Phi$ at the position $C$.

Then take derivative of $\|\nabla\Phi\|=1$, we have the following:
\begin{equation}\label{EQ:STB_NC3}
\nabla\|\nabla\Phi\|=(\nabla^2\Phi)\nabla\Phi=\vec{0},
\end{equation}
which completes the proof.
\end{proof}

\begin{proof}[Proof of Lemma \ref{TH:CEChain}]
We prove by induction.

For k=1, we have
\begin{equation}\label{EQ:CE_epsil_1}
{\partial C\over\partial t}=\epsilon(p,t)\vec{N}(p,t).
\end{equation}
Take derivative w.r.t t, we obtain the following:
\begin{equation}
{\partial^2 C\over\partial t^2} = {\partial\epsilon(p,t)\over\partial t}\vec{N}(p,t)+\epsilon(p,t){\partial\vec{N}\over \partial t}
={\partial\epsilon(p,t)\over\partial t}\vec{N}(p,t), \hbox{by lemma \ref{LM:DNN=0}}.
\end{equation}

Now we set $k=n$, we have
\begin{equation}
{\partial^{(n)} C\over\partial t^{(n)}}={\partial\epsilon^{n-1}(p,t)\over \partial t^{n-1}}\vec{N}(p,t).
\end{equation}
Hence, take derivative and by lemma \ref{LM:DNN=0},
\begin{equation}
{\partial^{(n+1)} C\over\partial t^{(n+1)}}={\partial\epsilon^{(n)}(p,t)\over\partial t^{(n)}}\vec{N}(p,t),
\end{equation}
which completes the proof.
\end{proof}

\begin{proof}[Proof of Lemma \ref{TH:DCE}]
The proof is constructive.

We expand $C(p,t+\tau)$ at $C(p,t)$ by Taylor series as follows:
\begin{equation}
C(p,t+\tau) = \sum\limits_{k=0}^{\infty}{\tau^k\over k!}{\partial^k C(p,t)\over\partial t^k}.
\end{equation}

By Lemma \ref{TH:CEChain}, we can rewrite the above as follows:
\begin{equation}
C(p,t+\tau)= C(p,t)+\left\{\sum\limits_{k=1}^{\infty}{\tau^k\over k!}{\partial^{(k-1)} \epsilon(p,t)\over\partial t^{(k-1)}}\right\}\vec{N}(p,t)=C(p,t)+\tau\xi(p,\tau)\vec{N}(p,t).
\end{equation}
Since the Taylor expansion exists, hence unique. By setting
\begin{equation}
\xi(p,\tau)=\sum\limits_{k=1}^{\infty}{\tau^{k-1}\over k!}{\partial^{(k-1)} \epsilon(p,t)\over\partial t^{(k-1)}},
\end{equation}
we complete the proof.
\end{proof}
\begin{proof}[Proof of Theorem \ref{TH:STB}]
We expect to obtain a differential equation that defines a contraction mapping, we then can show that the $\eta$ approaches $0$.

Firstly, according to Taylor's theorem, the curve evolution equation can be reformulated as follows:
\begin{equation}\label{EQ:STB1}
\left.{\partial C\over \partial t}\right|_{t=t^*} =\left.{\partial C\over \partial t}\right|_{t=t^*-\tau}+\tau\left.{\partial\over\partial t}{\partial C\over \partial t}\right|_{t=t^*-\tau}+O(\tau).
\end{equation}
This equation is the basic form for us to construct the contraction mapping. Our idea is to use $\eta$ to replace the $C$ in the equation to obtain our formulation.

Noting that
\begin{equation}
t=t^*\Leftrightarrow \tau=0,\hbox{and~~} {dt\over d\tau} = {d\tau\over dt}=-1.
\end{equation}
Differentiating Equation (\ref{EQ:STB0}), we can write the following:
\begin{equation}\label{EQ:STB1_}
{\partial\eta(p,\tau)\over\partial \tau}=-\left.{\partial C\over \partial t}\right|_{t=t^*-\tau}(-1).
\end{equation}

This relationship allow us to replace $C$ with $\eta$ for one of the terms of Equation (\ref{EQ:STB1}) to construct the contraction mapping defined by a differential equation.

Now we focus on replacing the $C$ with $\eta$ for the other term in Equation (\ref{EQ:STB1}). By making use of the Equation (\ref{EQ:DNN=0}), we have
\begin{equation}\label{EQ:STB2}
\begin{split}
&\left.{\partial\over\partial t}{\partial C\over \partial t}\right|_{t=t^*-\tau}=\left.{\partial\over\partial t}\langle\vec{F}(C),\vec{N}(C)\rangle\vec{N}(C)\right|_{t=t^*-\tau}\\
&=\left.\left(\left\langle J[\vec{F}](C){\partial C\over \partial t},\vec{N}(C)\right\rangle\vec{N}(C)
+\left\langle\vec{F}(C),\underbrace{{d\vec{N}\over dC}{\partial C\over \partial t}}\limits_{=0}\right\rangle\vec{N}(C)
+{\langle\vec{F}(C),\vec{N}(C)\rangle}\underbrace{{d\vec{N}\over dC}{\partial C\over \partial t}}\limits_{=0}\right)\right|_{t=t^*-\tau}\\
&=\left.[\vec{N}^TJ[\vec{F}]\vec{N}](C){\partial \eta\over\partial \tau}(C)\right|_{t=t^*-\tau}.
\end{split}
\end{equation}
where ${d\vec{N}\over dC}{\partial C\over \partial t}=0$  is by Lemma \ref{LM:DNN=0};

We now need the Taylor expansion of $\eta$ as follows.
\begin{equation}\label{EQ:STB_eta}
\eta(p,0)=\eta(p,\tau)-\tau{\partial \eta\over\partial \tau}+O(\tau)
\Leftrightarrow\eta(p,\tau)=\tau{\partial \eta\over\partial \tau}+O(\tau),
\end{equation}
where $\eta(p,0)=0$.

Substituting the above into Equation (\ref{EQ:STB2}), we obtain the following.
\begin{equation}\label{EQ:STB3}
\left.{\partial C\over \partial t}\right|_{t=t^*-\tau} = - \tau\left.{\partial\over\partial t}{\partial C\over \partial t}\right|_{t=t^*-\tau}+O(\tau)
= -[\vec{N}^TJ[\vec{F}]\vec{N}](C)\eta(p,\tau)+O(\tau).
\end{equation}

Combining the above with Equation (\ref{EQ:STB1_}), we finally arrive at the following,
\begin{equation}\label{EQ:STB5}
\begin{split}
\left.{\partial \eta\over \partial \tau}\right|_{\tau=t^*-t} &= -[\vec{N}^T(C)J[\vec{F}](C)\vec{N}(C)]\eta(p,\tau)
+O(\tau)
= -J_{nn}(C)\eta(p,\tau)+O(\tau),
\end{split}
\end{equation}
where
\begin{equation}\label{EQ:STB_Qnn}
\begin{split}
J_{nn}(C)=[\vec{N}^TJ[\vec{F}]\vec{N}](C).
\end{split}
\end{equation}

Now we almost obtain the differential equation in terms of $\eta$, but $C=C(p,t^*-\tau)$ is not convenient to deal with. We therefore use the following approximation:
\begin{equation}\label{EQ:STB_Qnn_Aprox}
J_{nn}(C)
=[\vec{N}^TJ[\vec{F}]\vec{N}](C^*)-\tau\left[\vec{N}^T[{dJ[\vec{F}]\over dC}\otimes{\partial C\over\partial t}]\vec{N}\right]+O(\tau)
\approx[\vec{N}^TJ[\vec{F}]\vec{N}](C^*),
\end{equation}
where ${dJ[\vec{F}]\over dC}$ is the second order derivative of the vector field, which is relatively small comparing to $J[\vec{F}]$. This treatment is valid for the case where the first order derivative does not vanish. Otherwise, it is categorized as the \emph{marginally stable}.

To further explain how the differential equation (\ref{EQ:STB5}) and the approximation Equation (\ref{EQ:STB_Qnn_Aprox}) can define a contraction mapping, and thus can be used for test of convergence, we discretize the differential equation and neglect the higher order terms $O(\tau)$ to obtain the following:
\begin{equation}\label{EQ:STB6}
\eta^k(p) = \eta^{k-1}(p)-_\Delta \tau J_{nn}(C^*)\eta^{k-1}(p)
       = \left(1-_\Delta \tau J_{nn}(C^*)\right)\eta^{k-1}(p).
\end{equation}

Then we take norm of the above to obtain the following:
\begin{equation}\label{EQ:STB7}
\|\eta^k(p)\|=\left|1-_\Delta \tau J_{nn}(C^*)\right|\|\eta^{k-1}(p)\|
=\left|1-_\Delta \tau J_{nn}(C^*)\right|^k\|\eta^{0}(p)\|.
\end{equation}

Noting that $\tau=t^*-t$ is decreasing, then $_\Delta \tau<0$. Hence, i.i.f $\left|1+_\Delta \tau J_{nn}(C^*)\right|<1$, $\lim\limits_{t\rightarrow\infty}\|\eta^{k}(p)\|=0$, i.e., $\eta$ will vanish. And i.i.f $J_{nn}(C^*)<0$, $\left|1+_\Delta \tau J_{nn}(C^*)\right|<1$.
\end{proof}

\begin{proof}[Proof of Lemma \ref{LM:OPT}]
Let us consider the function $\mathcal{G}(t)=g(C(p,t))$ for any fixed $p$, then by substituting in Equation (\ref{EQ:GVFEFa}), ${d \mathcal{G}(t)/ dt}=\|\langle\nabla g,\vec{N}\rangle\|^2\geq0$. Hence, $g$ is nondecreasing w.r.t $t$. Assuming $C(p,t^*)$ is the converged solution of Equation (\ref{EQ:GVFEFa}) where for all $\delta t>0$, $\left.\langle\nabla g,\vec{N}\rangle\right|_{t=t^*}=0$, we have that $g(C(p,t^*)) \geq g(C(p,t^*-\delta t))$.

By Theorem \ref{TH:DCE}, there exists $\xi(p)$ such that the following relation between $C(p,t^*-\delta t)$ and $C(p,t^*)$ holds:
\begin{equation}
C(p,t^*-\delta t)=C(p,t^*)+\xi(p)\vec{N}(C(p,t^*)).
\end{equation}

The Taylor expansion of $g$ at the converged solution $C(p,t^*)$, can be written as follows:
\begin{equation}\label{EQ:TLN1}
g(C(p,t^*-\delta t))
=g(C(p,t^*))+{\xi^2\over2}\left.\langle \nabla^2 g\vec{N},\vec{N}\rangle\right|_{t=t^*}
+\left.O(|\xi|^2)\right|_{t=t^*}.
\end{equation}
Since $g(C(p,t^*)) \geq g(C(p,t^*-\delta t))$, we have
\begin{equation}
\left.\langle \nabla^2 g\vec{N},\vec{N}\rangle\right|_{t=t^*}\leq0,
\end{equation}
which completes the proof.
\end{proof}